\documentclass[letterpaper, 10 pt, conference]{ieeeconf}  

\usepackage{booktabs}
\usepackage{graphicx} 
\usepackage{url}
\usepackage{hyperref}
\usepackage{caption}
\usepackage{marvosym}  
\usepackage{amsmath,amssymb}
\usepackage{algorithm}
\usepackage{algpseudocode}
\usepackage{url}
\IEEEoverridecommandlockouts                              

\title{\LARGE \bf
Multimodal Spatiotemporal-Frequency Fusion with Peak Enhancement for Cellular Traffic Forecasting
}
\author{Qingzhong Li$^{1}$, Yue Hu$^{1}$, Hui Ma$^{1,*}$, Yajun Zhang$^{1}$, Xinjun Pei$^{1}$, Ming Yan$^{1}$, and Fei Xing$^{2}$
\thanks{*Corresponding author: Hui Ma
        (e-mail: huima@xju.edu.cn)}%
\thanks{$^{1}$Qingzhong Li, Yue Hu, Hui Ma, Yajun Zhang, Xinjun Pei, and Ming Yan are with Xinjiang Key Laboratory of Intelligent Computing and Smart Applications,
        School of Software, Xinjiang University, Urumqi 830017, China.}%
\thanks{$^{2}$Fei Xing is with the College of Geography and Remote Sensing Sciences,
        Xinjiang University, Urumqi 830046, China.}%
}

\begin{document}

\maketitle

\thispagestyle{empty}
\pagestyle{empty}

\begin{abstract}

Accurate forecasting of cellular network traffic is essential for network planning, resource allocation, and quality-of-service assurance in modern mobile communication systems. Real-world traffic often exhibits bursty endogenous dynamics and disturbances triggered by external urban events, which makes reliable prediction highly challenging. Most existing spatiotemporal traffic forecasting methods primarily focus on intrinsic traffic patterns or structural relationships within a single modality, and rarely model burst behavior together with exogenous contextual signals. To address this issue, we propose \textbf{MSPF-Net}, a multimodal cellular traffic forecasting framework that integrates external contextual information. Specifically, MSPF-Net consists of a Spatiotemporal-Frequency Traffic Encoder for capturing temporal, spatial, and spectral traffic patterns, a Peak Enhancement Module for extracting burst-aware representations of sudden spikes, a News Context Representation Module for encoding urban news streams into exogenous contextual embeddings, and a Dynamic Fusion Prediction Module for adaptively integrating these heterogeneous signals to generate forecasts. Experiments on the Milano, Trento, and LTE traffic datasets demonstrate that jointly modeling traffic dynamics, burst patterns, and news contextual signals can effectively improve forecasting performance.

\end{abstract}

\section{INTRODUCTION}

Cellular networks constitute critical infrastructure for mobile communications and smart city services, and accurate traffic forecasting is essential for network planning, resource allocation, and quality-of-service assurance~\cite{Samudrala2026STAM,Pimpinella2026ContextAware}. Cellular traffic forecasting can be regarded as a spatiotemporal prediction task in complex sequential systems, requiring models to capture both temporal dependencies and spatial interactions. Although Transformer-based models and graph representation learning methods have demonstrated strong capability in modeling structured sequential data, real-world cellular traffic remains highly dynamic and is often influenced by bursty fluctuations and external events, making accurate forecasting still a challenging problem~\cite{Kougioumtzidis2025TFT}.

However, existing methods still face two major challenges. The first challenge lies in the inadequate modeling of burst-sensitive variations in real-world scenarios. Cellular traffic frequently exhibits sudden spikes, anomalies, and short-term fluctuations caused by event-driven demand surges or unexpected environmental changes. However, many existing methods either rely on unimodal traffic representations or adopt relatively shallow fusion strategies, making it difficult to characterize the complex interactions underlying these bursty dynamics~\cite{Chai2025UoMo}. As a result, current forecasting methods still show limitations in capturing local burst patterns and rapid traffic changes.

The second challenge lies in the insufficient utilization of exogenous contextual information~\cite{Alkadamani2026ContextualClustering}. Most existing methods mainly model intrinsic temporal or spatiotemporal dependencies based on historical traffic observations, assuming that the information required for prediction is largely contained in the observed traffic sequence itself. Although such designs are effective for learning regular traffic evolution patterns, they often underexplore heterogeneous external signals, such as urban events, news streams, and activity-related contextual cues, which can substantially influence traffic dynamics~\cite{Ma2023DSSMCellularTraffic}. Consequently, these methods may struggle to capture traffic changes triggered by external conditions that are not explicitly reflected in historical observations alone.

To address these challenges, we propose \textbf{MSPF-Net} (Multimodal Spatiotemporal-Frequency Fusion with Peak Enhancement Network), a unified multimodal cellular traffic forecasting framework that jointly models endogenous traffic dynamics and exogenous contextual signals. By integrating intrinsic spatiotemporal-frequency patterns, burst-aware representations, and contextual information derived from external events, MSPF-Net reformulates cellular traffic prediction as an adaptive multimodal fusion problem. The proposed framework consists of four cooperative components, including a traffic encoder, a peak enhancement module, a news context representation module, and a dynamic fusion prediction module. Through this coordinated design, MSPF-Net effectively captures endogenous traffic dynamics, burst-sensitive variations, and contextual influences, thereby enabling more robust forecasting under real-world disturbances.


We summarize the main contributions as follows:

\begin{itemize}
\item We propose a multimodal cellular traffic forecasting framework that integrates endogenous traffic dynamics with exogenous contextual signals, enabling joint modeling of intrinsic spatiotemporal-frequency patterns and external event influences.

\item We design a Peak Enhancement Module to learn burst-aware representations for capturing sudden traffic spikes and variation patterns that are often smoothed out by conventional sequence encoders.

\item We design a Dynamic Fusion Prediction Module to perform adaptive multimodal fusion of traffic representations, burst-aware representations, and news contextual signals, thereby improving forecasting performance compared with static fusion methods.
\end{itemize}

\section{RELATED WORK}

In recent years, research on cellular traffic prediction has gradually evolved from early univariate temporal modeling toward a more comprehensive paradigm that jointly captures multiple influencing factors and complex spatiotemporal couplings. A recent survey~\cite{Wang2024Survey} systematically reviews the progress of deep learning for cellular traffic prediction and indicates that existing methods can be broadly understood from four key perspectives: temporal dependencies, spatial dependencies, external factors and heterogeneity.

\subsection{Temporal Dependency Modeling}
Early studies on cellular traffic prediction mainly focused on modeling temporal dependencies from historical traffic observations, aiming to capture short-term fluctuations, long-range periodicity, and recurring temporal patterns in traffic sequences. For example, Shen \emph{et al.}~\cite{Shen2021TWACNet} further proposed TWACNet, which enhances the modeling of long-range temporal dependencies through a time-wise attention mechanism. More recent studies have attempted to further improve temporal representation learning. For instance, Riaz \emph{et al.}~\cite{Riaz2025Time2VecLSTM} introduced Time2Vec temporal embeddings to enhance the ability of LSTM to capture hourly and daily periodic patterns, while Samudrala and Senapati~\cite{Samudrala2025ATAM} proposed ATAM to emphasize critical sequential patterns via a temporal attention mechanism, thereby improving traffic prediction performance for 5G and beyond cellular networks. These methods have significantly improved the modeling of sequential dynamic features. However, relying solely on temporal modeling is still insufficient for cellular traffic prediction, since traffic evolution is also strongly influenced by spatial interactions among neighboring cells or regions.

\subsection{Spatial Dependency Modeling}

To better characterize spatial correlations in communication networks, subsequent studies have extended temporal modeling to spatiotemporal learning. For example, Weng \emph{et al.}~\cite{Weng2023DDGCRN} proposed DDGCRN, which introduces a signal decomposition mechanism into a dynamic graph convolutional recurrent network to separate normal and abnormal signals and enhance the modeling of complex spatiotemporal dependencies. Wang \emph{et al.}~\cite{Wang2023AHSTGNN} further learned adaptive hybrid spatial relations while additionally considering multi-periodic temporal inputs and cell-level heterogeneity. More recent studies have also begun to incorporate frequency-domain modeling into spatiotemporal learning. For instance, Li \emph{et al.}~\cite{Li2026GraFSTNet} proposed GraFSTNet, which combines graph-based spatial modeling with time-frequency analysis, while Teng \emph{et al.}~\cite{Teng2025FISTGCN} further proposed FISTGCN to enhance the representation of multi-scale spatial dependencies and complex traffic patterns through a frequency-aware and interactive spatiotemporal graph convolution mechanism. Nevertheless, many existing methods still mainly focus on intrinsic spatiotemporal traffic relations, while the modeling of burst dynamics under complex external perturbations remains limited.

\subsection{External-Factor and Heterogeneity-Aware Modeling} 
Besides endogenous traffic observations, cellular traffic is also influenced by external factors such as urban functionality, crowd activities, and event-driven disturbances, while real-world forecasting scenarios further involve substantial heterogeneity across regions, spatial structures, and contextual signals. To address these issues, a line of studies incorporates external variables and heterogeneous information into cellular traffic forecasting frameworks. For example, Assem \emph{et al.}~\cite{Assem2018STDenNetFus} integrated spatial, temporal, and external urban information for network demand prediction, while Zhang \emph{et al.}~\cite{Zhang2019STCNet} proposed a deep transfer learning framework to capture external factors affecting traffic generation. In addition, Yao \emph{et al.}~\cite{Yao2023MVSTGN} modeled heterogeneous spatial relations through multiple graph views. More recently, Ma and Yang~\cite{Ma2025MetaSTNet} explored a multimodal meta-learning framework for cellular traffic prediction under data-scarce settings. These studies show that external-factor-aware and heterogeneity-aware modeling can provide useful cues beyond historical traffic itself. Nevertheless, external information is still often incorporated in a relatively shallow manner, and most methods only address specific forms of heterogeneity, leaving the joint modeling of heterogeneous intrinsic dynamics and burst-sensitive exogenous context insufficiently explored.

Existing studies have demonstrated the effectiveness of temporal modeling, spatiotemporal learning, and external-context-aware representation learning for cellular traffic forecasting. However, most methods either model intrinsic traffic dynamics and external context separately or adopt coarse fusion strategies that fail to capture their complex interactions. In contrast, the proposed framework jointly models endogenous traffic dynamics and exogenous contextual signals through coordinated representation learning and adaptive multimodal fusion, providing a unified pipeline for forecasting burst-sensitive traffic behavior under external event influences.

\section{METHOD}

\begin{figure}[t]
  \centering
  \includegraphics[width=\columnwidth]{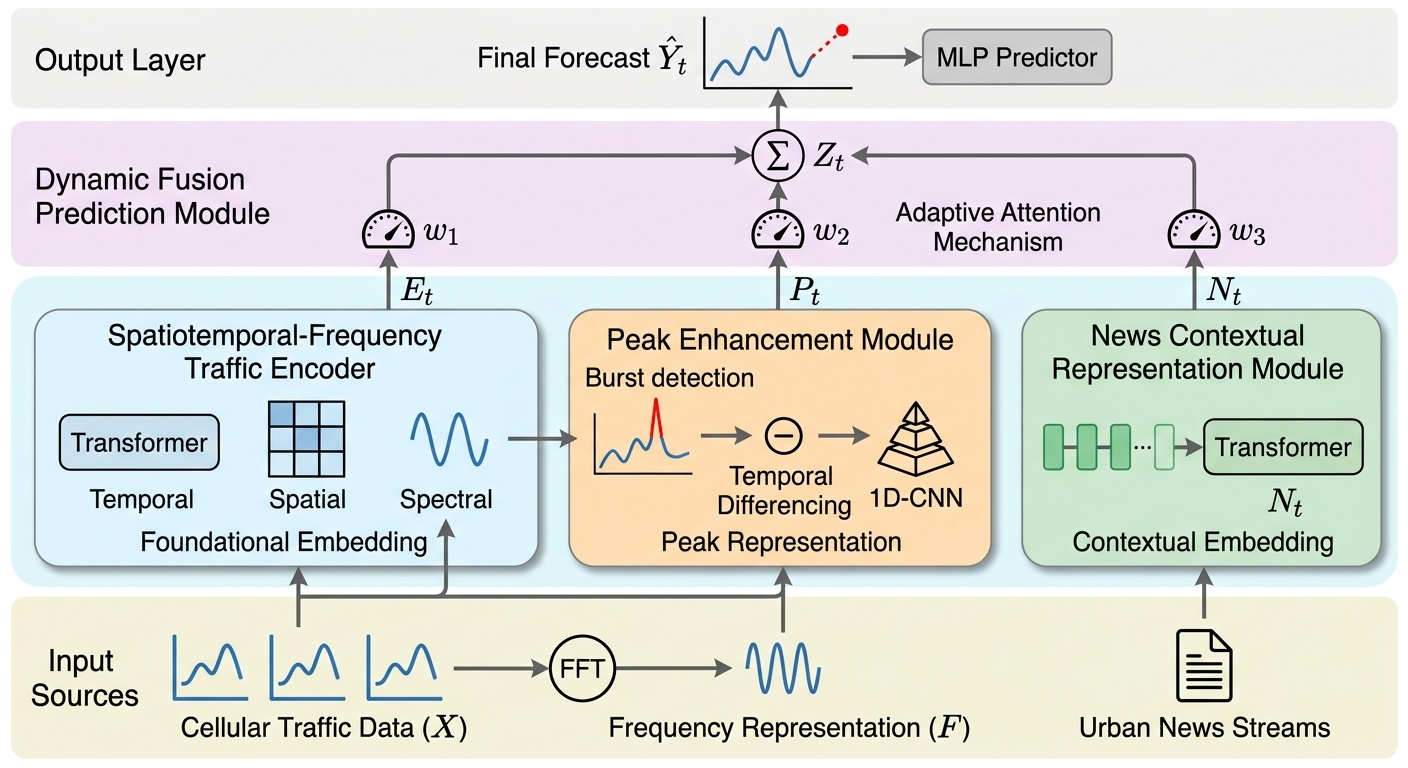}
  \caption{Overall architecture of the proposed multimodal spatio-temporal-frequency forecasting framework.}
  \label{fig:framework_overview}
\end{figure}

We propose \textbf{MSPF-Net}, a multimodal cellular traffic forecasting framework that jointly models endogenous traffic dynamics, burst-sensitive local variations, and exogenous news-driven context. MSPF-Net consists of four components: a \emph{Spatio-Temporal-Frequency Traffic Encoder}, a \emph{Peak Enhancement Module}, a \emph{News Context Representation Module}, and a \emph{Dynamic Fusion Prediction Module}. Given historical traffic observations, aligned news-derived contextual features, and the spatial graph of cellular regions, MSPF-Net first learns three complementary representations and then performs adaptive multimodal fusion to predict future traffic loads. The overall architecture is presented in Fig.~\ref{fig:framework_overview}.

\subsection{Problem Formulation}

Consider a cellular network with $N$ spatial units (e.g., base stations or spatial regions). Let $\mathbf{X}\in\mathbb{R}^{N\times L}$ denote the historical traffic sequence over the past $L$ time steps, where $\mathbf{X}_{i,t}$ is the traffic observation of node $i$ at time step $t$. Let $\mathbf{C}\in\mathbb{R}^{L\times F_c}$ denote the aligned news-derived contextual feature sequence, where $F_c$ is the contextual feature dimension. Let $\mathbf{A}\in\mathbb{R}^{N\times N}$ be the adjacency matrix describing spatial relations among cellular nodes. The forecasting objective is to predict the traffic values in the next $H$ time steps, denoted by $\hat{\mathbf{Y}}\in\mathbb{R}^{N\times H}$.

The overall forecasting function is defined as
\begin{equation}
\hat{\mathbf{Y}} = f_{\theta}(\mathbf{X}, \mathbf{C}, \mathbf{A}),
\label{eq:overall_mapping}
\end{equation}
where $f_{\theta}(\cdot)$ denotes the forecasting model with learnable parameters $\theta$.

\subsection{Spatio-Temporal-Frequency Traffic Encoder}






Cellular traffic exhibits coupled temporal, spatial, and spectral patterns. Temporal dependencies capture long-range traffic evolution, spatial dependencies describe interactions among neighboring nodes, and spectral patterns characterize periodic behaviors embedded in the traffic signal. To jointly model these characteristics, we construct a spatio-temporal-frequency traffic encoder.

We first project the raw traffic sequence into a latent space:
\begin{equation}
\mathbf{H}^{(0)} = \mathrm{Proj}_{x}(\mathbf{X}) + \mathbf{P}_{x},
\qquad
\mathbf{H}^{(0)} \in \mathbb{R}^{N \times L \times d},
\label{eq:input_projection}
\end{equation}
where $\mathrm{Proj}_{x}(\cdot)$ is a learnable linear projection, $\mathbf{P}_{x}$ is a positional embedding, and $d$ is the hidden dimension.

\paragraph{Temporal encoding}
To capture long-range temporal dependencies, we apply a temporal encoder to the projected traffic sequence:
\begin{equation}
\mathbf{H}^{(t)} = \mathrm{TempEnc}\!\left(\mathbf{H}^{(0)}\right),
\label{eq:temporal_encoder}
\end{equation}
where $\mathrm{TempEnc}(\cdot)$ is implemented using temporal self-attention followed by feed-forward transformation. This branch focuses on regular traffic evolution and temporal interactions across multiple time steps.

\paragraph{Frequency encoding}
To explicitly preserve periodic and oscillatory patterns, we transform the traffic representation into the frequency domain along the temporal axis:
\begin{equation}
\mathbf{F} = \left| \mathrm{FFT}\!\left(\mathbf{H}^{(t)}\right) \right|,
\label{eq:fft}
\end{equation}
where $\mathrm{FFT}(\cdot)$ denotes the fast Fourier transform and $|\cdot|$ extracts the amplitude spectrum. The spectral features are then encoded as
\begin{equation}
\mathbf{H}^{(f)} = \mathrm{FreqEnc}(\mathbf{F}),
\label{eq:frequency_encoder}
\end{equation}
where $\mathrm{FreqEnc}(\cdot)$ denotes the frequency encoder.

The temporal and frequency representations are aggregated through residual fusion:
\begin{equation}
\mathbf{H}^{(tf)} = \mathrm{LayerNorm}\!\left(\mathbf{H}^{(t)} + \mathbf{H}^{(f)}\right).
\label{eq:tf_fusion}
\end{equation}

\paragraph{Spatial encoding}
To model spatial correlations among cellular nodes, we perform graph-based propagation over the fused temporal-frequency representation. Let
\begin{equation}
\tilde{\mathbf{A}} = \mathbf{D}^{-\frac{1}{2}}(\mathbf{A} + \mathbf{I})\mathbf{D}^{-\frac{1}{2}}
\label{eq:normalized_adj}
\end{equation}
be the normalized adjacency matrix, where $\mathbf{D}$ is the degree matrix and $\mathbf{I}$ is the identity matrix. The spatial encoding is defined as
\begin{equation}
\mathbf{H}_{\mathrm{traf}} = \sigma\!\left( \tilde{\mathbf{A}} \, \mathbf{H}^{(tf)} \mathbf{W}_{s} \right),
\label{eq:spatial_encoder}
\end{equation}
where $\mathbf{W}_{s}$ is a learnable projection matrix and $\sigma(\cdot)$ is a nonlinear activation. The output $\mathbf{H}_{\mathrm{traf}} \in \mathbb{R}^{N \times L \times d}$ serves as the foundational traffic representation.
\subsection{Peak Enhancement Module}

\begin{figure}[t]
  \centering
  \includegraphics[width=\columnwidth]{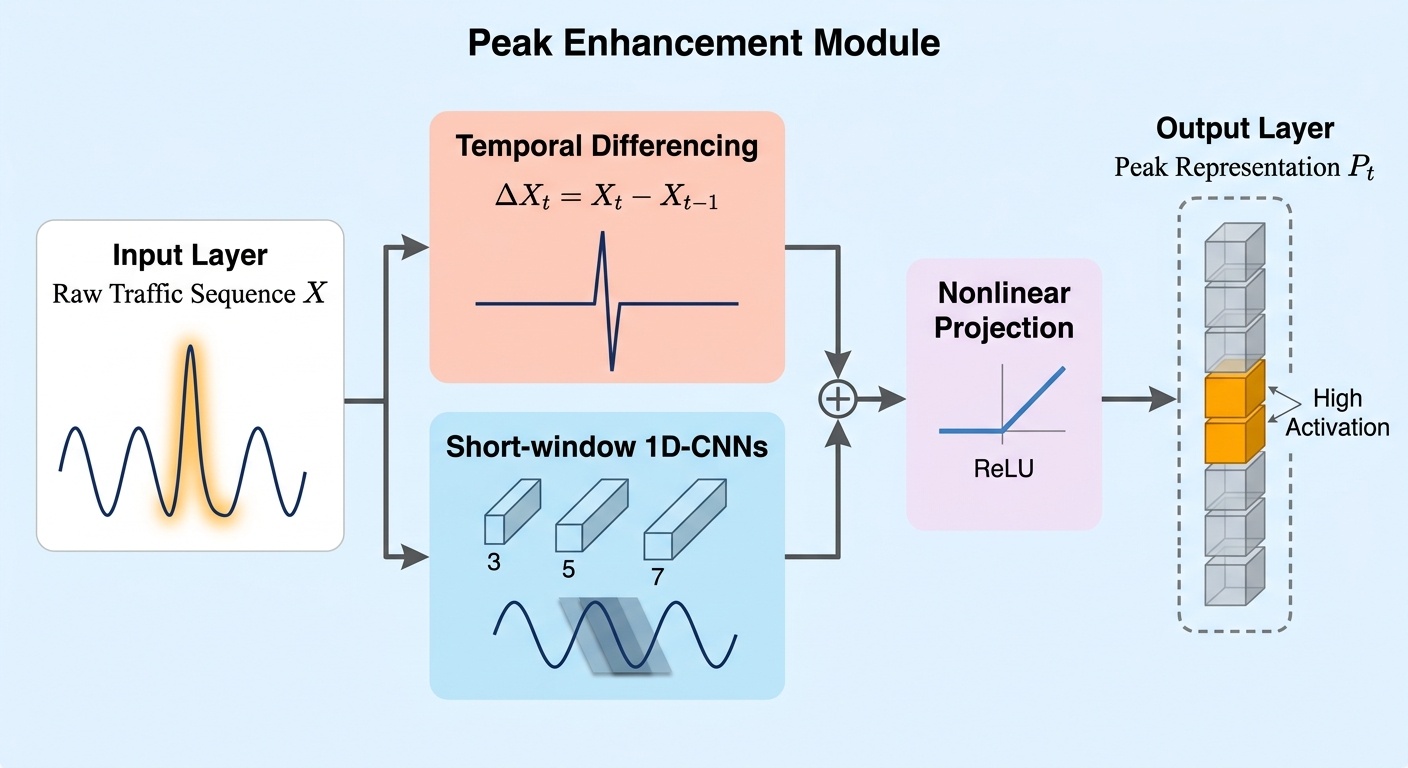}
  \caption{Peak enhancement module for modeling local abnormal spikes and mutation patterns in traffic sequences.}
  \label{fig:peak_module}
\end{figure}

Although the traffic encoder captures global spatio-temporal-frequency dynamics, bursty traffic spikes are typically sparse and can be smoothed out by dominant regular patterns. To explicitly preserve these high-impact local variations, we design a dedicated peak-aware branch, whose structure is shown in Fig.~\ref{fig:peak_module}.

We first compute the temporal difference to characterize short-term variation intensity, where $\Delta \mathbf{X}_{:,1}=\mathbf{0}$ and $\Delta \mathbf{X}_{:,t}=\mathbf{X}_{:,t}-\mathbf{X}_{:,t-1}$ for $t=2,\dots,L$.

To further describe local abnormal responses, we construct a peak descriptor using short-window statistics:
\begin{equation}
\mathbf{R}_{\mathrm{peak}} =
\mathrm{Concat}\!\Big(
\mathbf{X},
\Delta \mathbf{X},
\mathrm{MaxPool}_{w}(\mathbf{X}) - \mathrm{AvgPool}_{w}(\mathbf{X})
\Big),
\label{eq:peak_descriptor}
\end{equation}
where $\mathrm{MaxPool}_{w}(\cdot)$ and $\mathrm{AvgPool}_{w}(\cdot)$ denote max-pooling and average-pooling with window size $w$, respectively. The descriptor in \eqref{eq:peak_descriptor} jointly captures abrupt rises, local extreme responses, and peak-valley contrast.

The resulting descriptor is first projected into the hidden space and then encoded by a short-window 1D convolutional network:
\begin{equation}
\tilde{\mathbf{R}}_{\mathrm{peak}} = \mathrm{Proj}_{p}(\mathbf{R}_{\mathrm{peak}}),
\label{eq:peak_projection}
\end{equation}
\begin{equation}
\mathbf{U}_{\mathrm{peak}} = \mathrm{Conv1D}_{w_s}\!\left(\tilde{\mathbf{R}}_{\mathrm{peak}}\right),
\label{eq:peak_conv}
\end{equation}
where $\mathrm{Proj}_{p}(\cdot)$ is a learnable linear projection, $\mathrm{Conv1D}_{w_s}(\cdot)$ denotes a short-window 1D convolution with kernel size $w_s$ applied along the temporal dimension, and $\mathbf{U}_{\mathrm{peak}} \in \mathbb{R}^{N \times L \times d}$ is the intermediate peak feature map.

To enhance nonlinearity and obtain the final burst-aware representation, we further apply a nonlinear projection followed by a ReLU activation:
\begin{equation}
\mathbf{H}_{\mathrm{peak}} = \mathrm{ReLU}\!\left(\mathbf{U}_{\mathrm{peak}}\mathbf{W}_{p} + \mathbf{b}_{p}\right),
\label{eq:peak_encoder}
\end{equation}
where $\mathbf{W}_{p} \in \mathbb{R}^{d \times d}$ and $\mathbf{b}_{p} \in \mathbb{R}^{d}$ are learnable parameters. The output $\mathbf{H}_{\mathrm{peak}} \in \mathbb{R}^{N \times L \times d}$ emphasizes burst-sensitive local dynamics that are not sufficiently highlighted by the global traffic encoder.

\subsection{News Context Representation Module}
Cellular traffic is often affected by exogenous urban events, such as public gatherings, emergencies, and transportation disruptions. To encode such information, we construct an aligned news-derived contextual sequence from cleaned event statistics, which provides a compact description of external event intensity and activity patterns.

For each time step $t$, we define a five-dimensional context vector $\mathbf{c}_t \in \mathbb{R}^{5}$ as $\mathbf{c}_t = [u_t, s_t, e_t, r_t, m_t]$, where $u_t$, $s_t$, $e_t$, $r_t$, and $m_t$ denote the numbers of active users, mentioned cities, extracted entities, event types, and news items observed within interval $t$, respectively.

Traffic observations and contextual signals are organized at the same hourly resolution. For each hour, all cleaned news and event records within the corresponding interval are aggregated to construct the context vector $\mathbf{c}_t$; if no related record is observed, $\mathbf{c}_t$ is set to a zero vector. By stacking the context vectors from the past $L$ time steps, we obtain the aligned contextual feature sequence $\mathbf{C}\in\mathbb{R}^{L\times F_c}$, where $F_c=5$. To avoid information leakage, $\mathbf{C}$ is constructed only from records whose timestamps are not later than $t$, and all contextual dimensions are normalized using training-set statistics only.
Given the aligned contextual sequence $\mathbf{C}$, we first map it into the latent space:
\begin{equation}
\mathbf{E}_{\mathrm{news}} = \mathbf{C}\mathbf{W}_{c} + \mathbf{b}_{c} + \mathbf{P}_{c},
\qquad
\mathbf{E}_{\mathrm{news}} \in \mathbb{R}^{L \times d},
\label{eq:news_embedding}
\end{equation}
where $\mathbf{W}_{c} \in \mathbb{R}^{F_c \times d}$ and $\mathbf{b}_{c}$ are learnable parameters, and $\mathbf{P}_{c}$ is the positional encoding.

The temporal evolution of exogenous context is then modeled by a Transformer encoder:
\begin{equation}
\mathbf{H}_{\mathrm{news}}^{(c)} = \mathrm{TransformerEnc}\!\left(\mathbf{E}_{\mathrm{news}}\right),
\label{eq:news_context_encoder}
\end{equation}
where $\mathrm{TransformerEnc}(\cdot)$ consists of stacked self-attention and position-wise feed-forward blocks. The output $\mathbf{H}_{\mathrm{news}}^{(c)} \in \mathbb{R}^{L \times d}$ captures the temporal dynamics of external event signals.

To align the contextual representation with node-level traffic features, we broadcast it across the spatial dimension:
\begin{equation}
\mathbf{H}_{\mathrm{news}} = \mathbf{1}_{N} \otimes \mathbf{H}_{\mathrm{news}}^{(c)},
\label{eq:news_broadcast}
\end{equation}
where $\mathbf{1}_{N}$ denotes an all-one vector of length $N$, $\otimes$ denotes spatial replication, and $\mathbf{H}_{\mathrm{news}} \in \mathbb{R}^{N \times L \times d}$. In this way, each spatial node shares the same time-aligned exogenous context at each time step for subsequent fusion with endogenous traffic representations.

\subsection{Dynamic Fusion Prediction Module}
\begin{figure}[t]
  \centering
  \includegraphics[width=0.9\columnwidth]{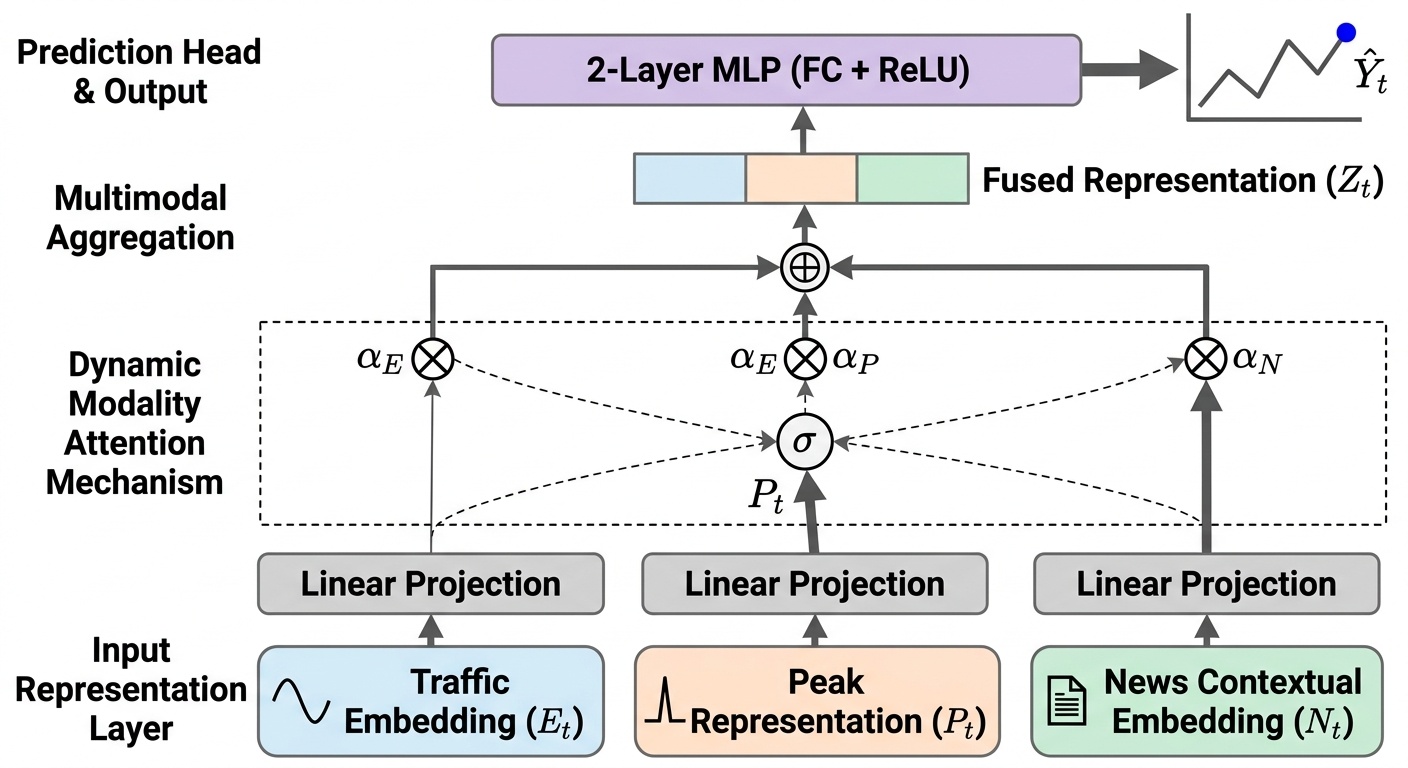}
  \caption{Dynamic multimodal fusion strategy that adaptively combines endogenous and exogenous representations.}
  \label{fig:dynamic_fusion}
\end{figure}

After obtaining $\mathbf{H}_{\mathrm{traf}}$, $\mathbf{H}_{\mathrm{peak}}$, and $\mathbf{H}_{\mathrm{news}}$, the model performs adaptive multimodal fusion to integrate endogenous traffic dynamics, burst-aware local variations, and exogenous contextual information. The fusion process is illustrated in Fig.~\ref{fig:dynamic_fusion}.

We first project the three representations into a shared latent space:
\begin{equation}
\begin{aligned}
\bar{\mathbf{H}}_{\mathrm{traf}} &= \mathbf{H}_{\mathrm{traf}}\mathbf{W}_{t}, \qquad
\bar{\mathbf{H}}_{\mathrm{peak}} = \mathbf{H}_{\mathrm{peak}}\mathbf{W}_{p}^{\prime}, \\
\bar{\mathbf{H}}_{\mathrm{news}} &= \mathbf{H}_{\mathrm{news}}\mathbf{W}_{n}.
\end{aligned}
\label{eq:shared_projection}
\end{equation}
where $\mathbf{W}_{t}$, $\mathbf{W}_{p}^{\prime}$, and $\mathbf{W}_{n}$ are learnable projection matrices.

We then use the traffic representation as the query and perform cross-modal interaction with the peak-aware and news-context branches:
\begin{equation}
\begin{aligned}
\mathbf{G}_{\mathrm{peak}}
&=
\mathrm{Attn}\!\left(
\bar{\mathbf{H}}_{\mathrm{traf}},
\bar{\mathbf{H}}_{\mathrm{peak}},
\bar{\mathbf{H}}_{\mathrm{peak}}
\right), \\
\mathbf{G}_{\mathrm{news}}
&=
\mathrm{Attn}\!\left(
\bar{\mathbf{H}}_{\mathrm{traf}},
\bar{\mathbf{H}}_{\mathrm{news}},
\bar{\mathbf{H}}_{\mathrm{news}}
\right).
\end{aligned}
\label{eq:cross_modal}
\end{equation}
The outputs $\mathbf{G}_{\mathrm{peak}}$ and $\mathbf{G}_{\mathrm{news}}$ enrich the traffic representation with burst-related cues and news-driven context, respectively.

To adaptively determine the contribution of each branch at each spatio-temporal position, we compute dynamic modality weights through a gating network:
\begin{equation}
[\alpha,\beta,\gamma] =
\mathrm{Softmax}\!\left(
\mathrm{MLP}\!\left(
\bar{\mathbf{H}}_{\mathrm{traf}} \,\|\, \mathbf{G}_{\mathrm{peak}} \,\|\, \mathbf{G}_{\mathrm{news}}
\right)
\right),
\label{eq:gating_weights}
\end{equation}
where $\alpha, \beta, \gamma \in \mathbb{R}^{N \times L \times 1}$ and $\|$ denotes concatenation along the feature dimension.

The fused multimodal representation is then obtained by
\begin{equation}
\mathbf{Z} =
\alpha \odot \bar{\mathbf{H}}_{\mathrm{traf}} +
\beta \odot \mathbf{G}_{\mathrm{peak}} +
\gamma \odot \mathbf{G}_{\mathrm{news}},
\label{eq:fused_representation}
\end{equation}
where $\odot$ denotes element-wise multiplication.

Next, a temporal pooling operation is applied over the historical horizon to aggregate the fused representation into a compact node-level summary.

Finally, the prediction head maps the aggregated representation to the forecasting horizon using a two-layer MLP:
\begin{equation}
\hat{\mathbf{Y}} =
\mathbf{W}_{o}^{(2)}
\,\mathrm{ReLU}\!\left(
\mathbf{m}\mathbf{W}_{o}^{(1)} + \mathbf{b}_{o}^{(1)}
\right)
+ \mathbf{b}_{o}^{(2)},
\label{eq:prediction_head}
\end{equation}
where $\mathbf{W}_{o}^{(1)}$, $\mathbf{W}_{o}^{(2)}$, $\mathbf{b}_{o}^{(1)}$, and $\mathbf{b}_{o}^{(2)}$ are learnable parameters, and $\hat{\mathbf{Y}} \in \mathbb{R}^{N \times H}$ is the final prediction.

\section{EXPERIMENTS}
\begin{table}[t]
\caption{Forecasting performance comparison on the Milano, Trento, and LTE traffic.}
\label{tab:main_results}
\centering
\small
\setlength{\tabcolsep}{2pt}
\begin{tabular}{lcc|cc|cc}
\toprule
\textbf{Method} 
& \multicolumn{2}{c|}{\textbf{Milano}} 
& \multicolumn{2}{c|}{\textbf{Trento}} 
& \multicolumn{2}{c}{\textbf{LTE traffic}} \\
& \textbf{MAE} & \textbf{RMSE} 
& \textbf{MAE} & \textbf{RMSE}
& \textbf{MAE} & \textbf{RMSE} \\
\midrule
LSTM                     & 5.5759 & 9.6049  & 6.3822 & 10.0130 & 0.9456 & 1.3363 \\
Transformer              & 4.9885 & 8.9165  & 5.7098 & 9.2954  & 0.8460 & 1.2406 \\
FEDformer~\cite{Zhou2022FEDformer} & 4.9028 & 8.7651  & 5.6117 & 9.1375  & 0.8314 & 1.2195 \\
TimeMixer~\cite{Wang2024TimeMixer} & 4.8168 & 8.6495  & 5.5132 & 9.0169  & 0.8169 & 1.2034 \\
\midrule
ST-Tran                  & 4.8029 & 8.5275  & 5.4974 & 8.8898  & 0.8145 & 1.1864 \\
DDGCRN~\cite{Weng2023DDGCRN} & 6.8104 & 12.4569 & 7.7951 & 12.9861 & 1.1549 & 1.7331 \\
OpenCity~\cite{Li2025OpenCity} & 5.1678 & 8.7731  & 5.9151 & 9.1458  & 0.8764 & 1.2206 \\
FISTGCN~\cite{Teng2025FISTGCN} & 8.3374 & 12.9921 & 9.5429 & 13.5442 & 1.4139 & 1.8076 \\
\midrule
MSCR~\cite{Althamary2024MSCR} & 3.9326 & 5.6841  & 3.4169 & 5.5361  & 0.5330 & 0.7152 \\
\textbf{MSPF-Net}        & \textbf{2.2850} & \textbf{3.7490} & \textbf{2.6154} & \textbf{3.9083} & \textbf{0.3875} & \textbf{0.5216} \\
\bottomrule
\end{tabular}
\end{table}
\subsection{Datasets}
\textbf{Milano.}
The Milano dataset\footnote{OpenStreet: \url{http://www.openstreet.com/}} was collected in Milan from November 1, 2013 to January 1, 2014. The city is divided into 10,000 grids, each covering about $235 \times 235$ m$^2$. We use hourly telecommunications traffic as the prediction target, with social pulse signals and daily news reports as textual context.

\textbf{Trento.}
The Trento dataset\footnotemark[1] was collected in the Trentino region and contains network traffic records and text-based information, including social pulse signals and web news. We use the hourly traffic series for forecasting.

\textbf{LTE traffic.}
The LTE traffic dataset comes from a private telecom operator and contains two weeks of hourly downlink measurements. We use downlink traffic volume, downlink throughput, and temporal features as auxiliary inputs. Spatial visual context is obtained from OpenStreetMap\footnotemark[1], including map images and geographic coordinates.




\subsection{Overall Comparison}

\begin{figure}[t]
  \centering
  \includegraphics[width=0.95\columnwidth]{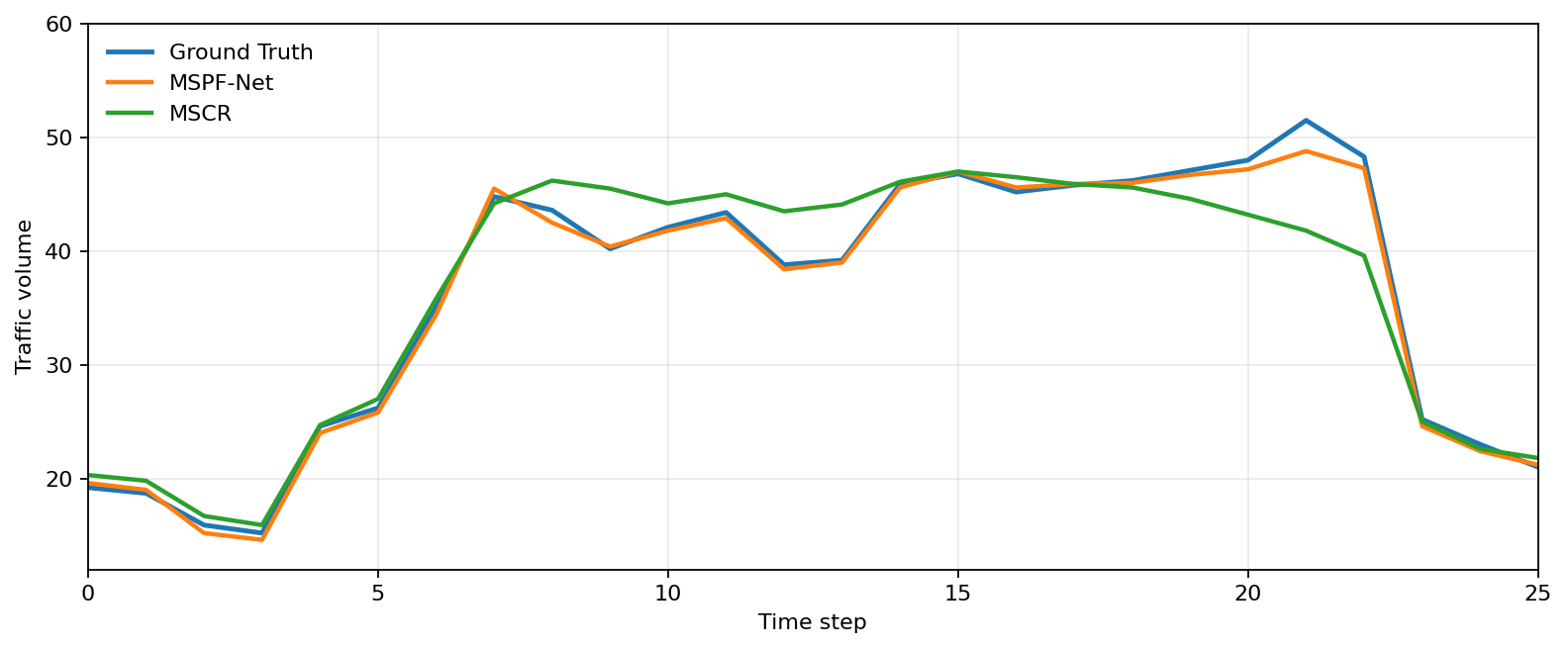}
  \includegraphics[width=0.95\columnwidth]{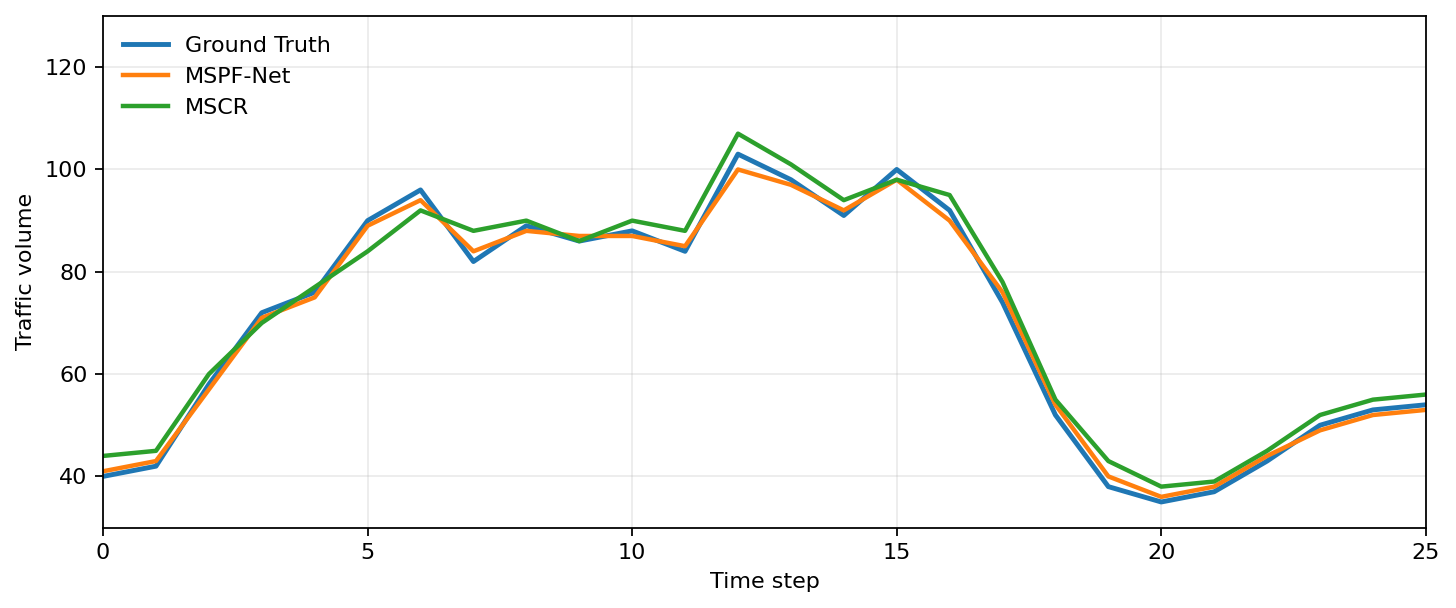}
   \includegraphics[width=0.95\columnwidth]{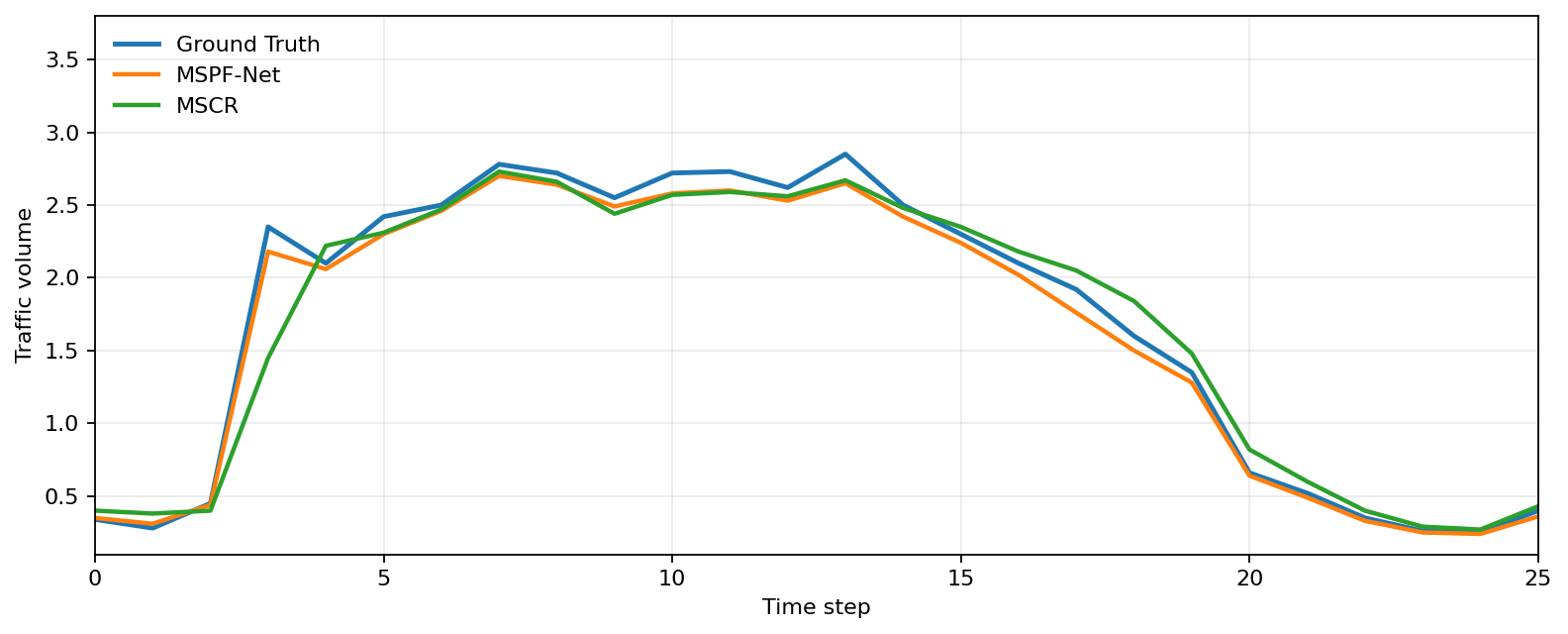}
  \caption{Comparison between the ground-truth values and the predicted values on the Milano ,Trento and LTE traffic datasets.}
  \label{fig:prediction_case}
\end{figure}

Table~\ref{tab:main_results} presents the forecasting results on the Milano, Trento, and LTE traffic datasets. LSTM improves prediction accuracy by modeling temporal dependencies, while Transformer enhances the modeling of long-range temporal dependencies through attention-based sequence representations. FEDformer further improves the characterization of global trends and periodic patterns in traffic sequences by explicitly introducing frequency-domain decomposition. TimeMixer strengthens the modeling of multi-scale temporal patterns. However, these models remain limited in their ability to capture complex spatial correlations. Graph-based methods, including DDGCRN, ST-Tran, OpenCity, and FISTGCN, reduce prediction errors by explicitly modeling spatial interactions among cellular regions. In addition, the results of MSCR indicate that incorporating contextual signals can further improve forecasting performance compared with unimodal methods. By contrast, the proposed multimodal cellular traffic forecasting framework achieves the lowest prediction errors on both datasets, demonstrating that jointly modeling endogenous traffic dynamics, burst-aware representations, and exogenous contextual signals leads to more accurate forecasts. Figure~\ref{fig:prediction_case} compares the ground-truth values and the predicted values on the Milano, Trento, and LTE traffic datasets, showing that MSPF-Net can more closely follow the real traffic trends and better capture sudden fluctuations than the second-best baseline.

\subsection{Ablation Study}
\begin{table}[t]
\caption{Ablation study on different components of MSPF-Net.}
\label{tab:ablation}
\centering
\small
\setlength{\tabcolsep}{0.8pt}
\begin{tabular}{lcc|cc|cc}
\toprule
\textbf{Method} & \multicolumn{2}{c|}{\textbf{Milano}} & \multicolumn{2}{c|}{\textbf{Trento}} & \multicolumn{2}{c}{\textbf{LTE traffic}} \\
 & \textbf{MAE} & \textbf{RMSE} & \textbf{MAE} & \textbf{RMSE} & \textbf{MAE} & \textbf{RMSE} \\
\midrule
\textbf{MSPF-Net}      & \textbf{2.2850} & \textbf{3.7490} & \textbf{2.6154} & \textbf{3.9083} & \textbf{0.3875} & \textbf{0.5216} \\
w/o News Context       & 2.4786          & 4.0584          & 2.8567          & 4.3271          & 0.4218          & 0.5711          \\
w/o Peak Enhancement   & 2.5348          & 4.1267          & 2.9102          & 4.3985          & 0.4305          & 0.5806          \\
w/o Dynamic Fusion     & 2.6129          & 4.2842          & 3.0284          & 4.6156          & 0.4461          & 0.6060          \\
\bottomrule
\end{tabular}
\end{table}
Table~\ref{tab:ablation} reports the contribution of each component. Removing the Peak Enhancement Module causes a clear accuracy drop, especially during burst-heavy intervals, because short-term spikes are smoothed by global traffic and contextual representations. Removing the News Contextual Representation Module also degrades performance, indicating that endogenous traffic signals alone are insufficient to handle exogenous disturbances reflected in news events. Replacing the Dynamic Fusion Prediction Module with static aggregation further reduces accuracy on both datasets, since fixed fusion cannot adapt the importance of traffic, burst, and contextual cues under changing conditions. These results verify that peak-aware modeling and adaptive multimodal fusion are both critical for robust forecasting.

\section{CONCLUSION}

This study investigates cellular traffic forecasting from the perspective of multimodal spatiotemporal predictive modeling and proposes a unified multimodal forecasting framework, namely MSPF-Net. The framework consists of a Spatio-Temporal-Frequency Traffic Encoder, a Peak Enhancement Module, a News Context Representation Module, and a Dynamic Fusion Prediction Module, which are designed to jointly model endogenous traffic dynamics and exogenous contextual signals. Experimental results on the Milano, Trento, and LTE traffic datasets show that, by adaptively fusing traffic embeddings, burst-aware representations, and contextual embeddings, MSPF-Net achieves better forecasting performance than other baseline methods, especially in burst-heavy and event-driven scenarios. These results indicate that treating cellular traffic forecasting as a multimodal representation learning problem helps more comprehensively capture spatiotemporal dependencies, spectral periodic patterns, and the influence of external events. Future work may explore richer graph-based modeling of spatial interactions and more expressive sequence modeling strategies, while also improving the alignment between traffic signals and real-world event streams, so as to further enhance multimodal traffic forecasting.

\textbf{Acknowledgments.} 
This work was supported by the Tianchi Talents - Young Doctor Program (5105250183m),  Science and Technology Program of Xinjiang Uyghur Autonomous Region (2025B04051, 2024B03028), Regional Fund of the National Natural Science Foundation of China (202512120005).




\bibliographystyle{IEEEtran}   
\bibliography{mybibfile}      

\begin{thebibliography}{10}
\providecommand{\url}[1]{#1}
\csname url@samestyle\endcsname
\providecommand{\newblock}{\relax}
\providecommand{\bibinfo}[2]{#2}
\providecommand{\BIBentrySTDinterwordspacing}{\spaceskip=0pt\relax}
\providecommand{\BIBentryALTinterwordstretchfactor}{4}
\providecommand{\BIBentryALTinterwordspacing}{\spaceskip=\fontdimen2\font plus
\BIBentryALTinterwordstretchfactor\fontdimen3\font minus
  \fontdimen4\font\relax}
\providecommand{\BIBforeignlanguage}[2]{{%
\expandafter\ifx\csname l@#1\endcsname\relax
\typeout{** WARNING: IEEEtran.bst: No hyphenation pattern has been}%
\typeout{** loaded for the language `#1'. Using the pattern for}%
\typeout{** the default language instead.}%
\else
\language=\csname l@#1\endcsname
\fi
#2}}
\providecommand{\BIBdecl}{\relax}
\BIBdecl

\bibitem{Samudrala2026STAM}
D.~S. Samudrala and R.~Senapati, ``Spatio temporal attention mechanism for real
  time cellular traffic prediction,'' \emph{PeerJ Computer Science}, vol.~12,
  p. e3571, 2026.

\bibitem{Pimpinella2026ContextAware}
A.~Pimpinella and A.~E.~C. Redondi, ``Generative-aided and context-aware
  forecasting of mobile network traffic,'' \emph{Computer Networks}, p. 112147,
  2026, available online 27 March 2026, journal pre-proof.

\bibitem{Kougioumtzidis2025TFT}
G.~Kougioumtzidis, V.~K. Poulkov, P.~I. Lazaridis, and Z.~D. Zaharis, ``Mobile
  network traffic prediction using temporal fusion transformer,'' \emph{IEEE
  Transactions on Artificial Intelligence}, vol.~6, no.~10, pp. 2685--2699,
  2025.

\bibitem{Chai2025UoMo}
H.~Chai, S.~Zhang, X.~Qi, B.~Qiu, and Y.~Li, ``Uomo: A universal model of
  mobile traffic forecasting for wireless network optimization,'' in
  \emph{Proceedings of the 31st ACM SIGKDD Conference on Knowledge Discovery
  and Data Mining}, 2025.

\bibitem{Alkadamani2026ContextualClustering}
M.~Alkadamani, C.~Brown, and H.~Yanikomeroglu, ``{AI-Enhanced Spatial Cellular
  Traffic Demand Prediction with Contextual Clustering and Error Correction for
  5G/6G Planning},'' \emph{arXiv preprint arXiv:2603.10800}, 2026.

\bibitem{Ma2023DSSMCellularTraffic}
H.~Ma, K.~Yang, and M.~O. Pun, ``Cellular traffic prediction via deep state
  space models with attention mechanism,'' \emph{Computer Communications}, vol.
  197, pp. 276--283, 2023.

\bibitem{Wang2024Survey}
X.~Wang, Z.~Wang, K.~Yang, Z.~Song, C.~Bian, J.~Feng, and C.~Deng, ``A survey
  on deep learning for cellular traffic prediction,'' \emph{Intelligent
  Computing}, vol.~3, p. 0054, 2024.

\bibitem{Shen2021TWACNet}
W.~Shen, H.~Zhang, S.~Guo, and C.~Zhang, ``Time-wise attention aided
  convolutional neural network for data-driven cellular traffic prediction,''
  \emph{IEEE Wireless Communications Letters}, vol.~10, no.~8, pp. 1747--1751,
  2021.

\bibitem{Riaz2025Time2VecLSTM}
H.~Riaz, S.~{\"O}zt{\"u}rk, and P.~G{\"u}ne{\c{s}}, ``Improving cellular
  traffic prediction with temporal embeddings: A time2vec-lstm approach,''
  \emph{International Journal of Engineering Technologies IJET}, vol.~10,
  no.~2, pp. 48--56, Sep. 2025.

\bibitem{Samudrala2025ATAM}
D.~S. Samudrala and R.~Senapati, ``{Advanced Temporal Attention Mechanism Based
  Traffic Prediction Model for 5G and Beyond Cellular Networks},'' \emph{ICT
  Express}, 2025.

\bibitem{Weng2023DDGCRN}
W.~Weng, J.~Fan, H.~Wu, Y.~Hu, H.~Tian, F.~Zhu, and J.~Wu, ``A decomposition
  dynamic graph convolutional recurrent network for traffic forecasting,''
  \emph{Pattern Recognition}, vol. 142, p. 109670, 2023.

\bibitem{Wang2023AHSTGNN}
X.~Wang, K.~Yang, Z.~Wang, J.~Feng, L.~Zhu, J.~Zhao, and C.~Deng, ``Adaptive
  hybrid spatial-temporal graph neural network for cellular traffic
  prediction,'' in \emph{ICC 2023 - IEEE International Conference on
  Communications}, 2023, pp. 4026--4032.

\bibitem{Li2026GraFSTNet}
Z.~Li, H.~Ma, F.~Xing, C.~Zhang, and M.~Yan, ``Grafstnet: Graph-based frequency
  spatiotemporal network for cellular traffic prediction,'' \emph{arXiv
  preprint arXiv:2602.13282}, 2026.

\bibitem{Teng2025FISTGCN}
G.~Teng, H.~Wu, H.~Wu, J.~Cao, and M.~Zhao, ``Frequency-aware and interactive
  spatial-temporal graph convolutional network for traffic flow prediction,''
  \emph{Applied Sciences}, vol.~15, no.~20, p. 11254, 2025.

\bibitem{Assem2018STDenNetFus}
H.~Assem, B.~Caglayan, T.~S. Buda, and D.~O'Sullivan, ``St-dennetfus: A new
  deep learning approach for network demand prediction,'' in \emph{Machine
  Learning and Knowledge Discovery in Databases}, ser. Lecture Notes in
  Computer Science.\hskip 1em plus 0.5em minus 0.4em\relax Springer, 2018, pp.
  222--237.

\bibitem{Zhang2019STCNet}
C.~Zhang, H.~Zhang, J.~Qiao, D.~Yuan, and M.~Zhang, ``Deep transfer learning
  for intelligent cellular traffic prediction based on cross-domain big data,''
  \emph{IEEE Journal on Selected Areas in Communications}, vol.~37, no.~6, pp.
  1389--1401, 2019.

\bibitem{Yao2023MVSTGN}
Y.~Yao, B.~Gu, Z.~Su, and M.~Guizani, ``{MVSTGN: A Multi-View Spatial-Temporal
  Graph Network for Cellular Traffic Prediction},'' \emph{IEEE Transactions on
  Mobile Computing}, vol.~22, no.~5, pp. 2837--2849, 2021.

\bibitem{Ma2025MetaSTNet}
H.~Ma and K.~Yang, ``{MetaSTNet: Multimodal Meta-Learning for Cellular Traffic
  Conformal Prediction},'' \emph{IEEE Transactions on Network Science and
  Engineering}, vol.~11, no.~2, pp. 1999--2011, 2023.

\bibitem{Zhou2022FEDformer}
T.~Zhou, Z.~Ma, Q.~Wen, X.~Wang, L.~Sun, and R.~Jin, ``Fedformer: Frequency
  enhanced decomposed transformer for long-term series forecasting,'' in
  \emph{Proceedings of the 39th International Conference on Machine Learning},
  ser. Proceedings of Machine Learning Research, vol. 162.\hskip 1em plus 0.5em
  minus 0.4em\relax PMLR, 2022, pp. 27\,268--27\,286.

\bibitem{Wang2024TimeMixer}
S.~Wang, H.~Wu, X.~Shi, T.~Hu, H.~Luo, L.~Ma, J.~Y. Zhang, and J.~Zhou,
  ``Timemixer: Decomposable multiscale mixing for time series forecasting,'' in
  \emph{The Twelfth International Conference on Learning Representations},
  2024.

\bibitem{Li2025OpenCity}
Z.~Li, L.~Xia, L.~Shi, Y.~Xu, D.~Yin, and C.~Huang, ``Opencity: Open
  spatio-temporal foundation models for traffic prediction,'' \emph{ACM
  Transactions on Intelligent Systems and Technology}, 2025.

\bibitem{Althamary2024MSCR}
I.~A. Althamary, R.~Boisguene, and C.-W. Huang, ``Enhanced multi-task traffic
  forecasting in beyond 5g networks: Leveraging transformer technology and
  multi-source data fusion,'' \emph{Future Internet}, vol.~16, no.~5, p. 159,
  2024.

\end{thebibliography}

\end{document}